\documentclass[tinyml]{acmart}
\usepackage{url, multirow, makecell, framed, mdframed, enumitem, fancyvrb, amsmath, graphicx, color}
\newcommand{\eqdef}{\overset{\mathrm{def}}{=\joinrel=}}

\AtBeginDocument{%
  \providecommand\BibTeX{{%
    \normalfont B\kern-0.5em{\scshape i\kern-0.25em b}\kern-0.8em\TeX}}}

\setcopyright{rightsretained}
\copyrightyear{2022}
\acmYear{2022}

\begin{document}

\title[PocketNN: Integer-only Training and Inference of Neural Networks via DFA and Pocket Activations in Pure C++]
{PocketNN: Integer-only Training and Inference of Neural Networks via Direct Feedback Alignment and Pocket Activations in Pure C++}

\author{Jaewoo Song}
\affiliation{%
  \institution{Department of Computer Science and Engineering \protect\\ The Hong Kong University of Science and Technology}
  \streetaddress{Clear Water Bay}
  \city{Kowloon}
  \state{Hong Kong}
  \country{China}
}
\email{jsongab@connect.ust.hk}

\author{Fangzhen Lin}
\affiliation{%
  \institution{Department of Computer Science and Engineering \protect\\ The Hong Kong University of Science and Technology}
  \streetaddress{Clear Water Bay}
  \city{Kowloon}
  \state{Hong Kong}
  \country{China}
}
\email{flin@cse.ust.hk}


\begin{abstract}
Standard deep learning algorithms are implemented using floating-point real numbers. This presents an obstacle for implementing them on low-end devices which may not have dedicated floating-point units (FPUs). As a result, researchers in tinyML have considered machine learning algorithms that can train and run a deep neural network (DNN) on a low-end device using integer operations only. In this paper we propose PocketNN, a light and self-contained proof-of-concept framework in pure C++ for the training and inference of DNNs using only integers. Unlike other approaches, PocketNN directly operates on integers without requiring any explicit quantization algorithms or customized fixed-point formats. This was made possible by pocket activations, which are a family of activation functions devised for integer-only DNNs, and an emerging DNN training algorithm called direct feedback alignment (DFA). Unlike the standard backpropagation (BP), DFA trains each layer independently, thus avoiding integer overflow which is a key problem when using BP with integer-only operations. We used PocketNN to train some DNNs on two well-known datasets, MNIST and Fashion-MNIST. Our experiments show that the DNNs trained with our PocketNN achieved 96.98\% and 87.7\% accuracies on MNIST and Fashion-MNIST datasets, respectively. The accuracies are very close to the equivalent DNNs trained using BP with floating-point real number operations, such that accuracy degradations were just 1.02\%p and 2.09\%p, respectively. Finally, our PocketNN has high compatibility and portability for low-end devices as it is open source and implemented in pure C++ without any dependencies.
\end{abstract}

\keywords{PocketNN, Pocket Activations, tinyML, deep learning, deep neural network, DNN, integer, integer-only, training, inference, direct feedback alignment, DFA, quantization, C++, framework}
\maketitle

\section{Introduction}
Deep learning has shown great advances in recent years, and it is now widely used in academia and industry. However, big size of recent deep neural network (DNN) models made it very hard to run them, and even harder to train them, without high-end hardware such as fast GPUs and large memory. This is a big problem because low-end devices are becoming more and more prevalent in many areas of the real world and so does machine learning for those devices \cite{tinyml2021prospect}.

TinyML is the type of machine learning for low-end devices. Because of hardware limitations, current practice of tinyML is to run only simple models on low-end devices. If a complex inference is needed, low-end devices send input data to their servers and receive inference results. This causes higher electric power usage, longer time delay for online communication, and more expensive device price because of additional semiconductor chips for communication modules. Training is another big problem. Since training cannot be performed on low-end devices, training data need to be sent to servers. Including all the previously mentioned overhead, data privacy also becomes a big issue. Therefore, it will be very useful if complex DNN models can be run and trained on low-end devices. It will make low-end devices cheaper, consume less electric power, and solve the training data privacy problem.

Speed and size are the two main obstacles of running and training DNN models on low-end devices. Because low-end devices are slow and their storage is tiny, it is important to make DNN models faster and smaller. Quantization is a popular approach for that purpose. Quantization is the process of constraining parameters from many-bit floating-point real numbers to few-bit integers. Quantization makes not only model size smaller but also model inference process faster because integer operations are often much faster than floating-point operations on low-end devices. It is because many microcontrollers (MCUs) widely used in low-end devices do not have floating-point units (FPUs) and therefore floating-point arithmetic need to be emulated by software with severe slowdown. For example, Arduino Uno, Zero and Due do not have FPUs \cite{atmega328p} \cite{atsamd21g18a} \cite{sam3x8e}. Consequently, Arduino official website mentions ``\textit{Floating point math is also much slower than integer math in performing calculations, so should be avoided...}'' \cite{arduino2021}.

Previous research projects on quantization can be classified into two categories. The first category corresponds to quantizing inference only. BinaryConnect \cite{courbariaux2015binaryconnect}, Courbariaux et al. \cite{courbariaux2016binarized}, XNOR-net \cite{rastegari2016xnor}, TTQ \cite{zhu2016trained}, Benoit Jacob et al. \cite{jacob2018quantization}, and Ron Banner et al. \cite{banner2018scalable} belong to this category. They use high precision floating-point real numbers for training and weight update. Then the real-valued weights are quantized for inference. The strength of this approach is that models can be trained and quantized in high-end servers with big data. However, once the quantized models are delivered to the low-end devices, it is impossible to fine-tune the models locally. New models again need to be trained on the servers and sent to the low-end devices for update. Training data also need to be sent to the servers from the low-end devices. It should be noted that both quantization aware training (QAT) and post training quantization (PTQ), two popular DNN quantization methods, belong here because they use real numbers in their quantization processes.

The second category corresponds to quantizing both training and inference. This approach leverages advantages of the first category and enables training with integer arithmetic, which is faster and more compatible than floating-point arithmetic on many low-end devices. DoReFa-Net \cite{zhou2016dorefa}, FxpNet \cite{chen2017fxpnet}, WAGE \cite{wu2018training}, and NITI \cite{wang2020niti} belong to this category. Research of this category highly focuses on designing more accurate quantization algorithms with techniques such as bit shifting, scaling, deterministic and stochastic rounding methods, and customized fixed-point notations for simulating floating-point real numbers. However, those operations make the overall process complex. It should be noted that some research projects in this category still partially use floating-point formats \cite{zhou2016dorefa} \cite{chen2017fxpnet}. Customized fixed-point notations are also often used to simulate floating-point real numbers. For example, FxpNet \cite{chen2017fxpnet} and NITI \cite{wang2020niti} use two different integers $M$ and $E$ to express a single real number $M \cdot 2^E$. Essentially this is same as using two integers to simulate one floating-point real number of IEEE 754 format \cite{ieee754}. Therefore, although integers are used, \textit{conceptually} they are still using real numbers. This and other customized fixed-point formats are not universally agreed and therefore they suffer from low portability and compatibility. Additionally, many operations widely used in DNN training such as logarithms and square roots are hard to be efficiently implemented with integer-only operations.

Overflow is another big problem of this category. FxpNet \cite{chen2017fxpnet} and NITI \cite{wang2020niti} explicitly mentioned about overflow. We mathematically figured out that backpropagation (BP) \cite{rumelhart1986learning}, a de facto standard algorithm for DNN training, can very easily lead to overflow under integer-only arithmetic. This is because BP calculates gradients for weight parameter updates by propagating error backward via chain rule (hence ``backpropagation''). So upper bounds of the size of weight update grow exponentially layer by layer in BP under integer-only arithmetic. For example, if error and weights are $e$- and $w$-bit integers respectively, the upper bound of updates for weights of the $k$th hidden layer from the output becomes $O(2^{e+kw})$. So overflow will eventually occur as $k$ increases. For instance, if weights are represented in \texttt{int8} (i.e., $w = 8$), fourth hidden layer from the output ($k = 4$) can suffer from overflow under 32-bit integer-only arithmetic because $O(2^{e+kw}) = O(2^{e+32})$. Floating-point DNNs are relatively safe from this problem because floating-point arithmetic allows bits to be used to finely divide a small range instead of increasing the range.

Last but not least, all open source code from previous works were written in Python with popular deep learning libraries such as Tensorflow \cite{tensorflow2015-whitepaper} and PyTorch \cite{NEURIPS2019_9015}. It is hard to directly apply Python code to low-end devices because they usually support only C, C++ or assembly languages. Dependencies on other deep learning libraries make it even harder to port the algorithms to low-end devices.

PocketNN is an integer-only algorithm to solve the abovementioned problems, and also a name of the proof-of-concept C++ framework for that algorithm. As a framework, it is light and written in pure C++ without any external libraries to ensure maximum compatibility. As an algorithm, PocketNN uses only standard type integers, not floating-points nor any customized fixed-point formats. PocketNN directly operates on integers and quantization is naturally done without any explicit quantization algorithms. This was made possible by using direct feedback alignment \cite{nokland2016direct} instead of BP, together with specially designed integer activation functions called pocket activations.

Direct feedback alignment (DFA) \cite{nokland2016direct} is a new emerging DNN training algorithm. DFA has novel characteristic that random feedback weights can be used to train DNNs. Therefore weight updates for each layer can be calculated independently from weights of higher layers. This solves the overflow problem and also opens the potential for much simpler and easily parallelizable training. Combined with pocket activations, this is the first research paper of training DNNs entirely via DFA with integer-only arithmetic to the best of our knowledge.

To compare PocketNN to floating-point BP, fully connected neural networks were trained on the MNIST \cite{lecun1998gradient} and Fashion-MNIST \cite{fashionmnist2017} datasets. The PocketNN models achieved 96.98\% and 87.7\% accuracies on validation sets of the datasets and showed marginal accuracy degradations of 1.02\%p and 2.09\%p compared to floating-point BP, respectively.

PocketNN will be useful for tinyML researchers and developers who work on the training and inference of DNN models on low-end devices. PocketNN is open source and it can be downloaded at its online repository. \footnote{\urlstyle{tt}\url{https://github.com/jaewoosong/pocketnn}} In summary, the contributions of this paper are:
\begin{itemize}[leftmargin=*]
\item PocketNN, an algorithm for fully integer-only inference and training of DNNs via direct feedback alignment (DFA) and pocket activations.
\item Removal of complex quantization algorithms and customized fixed-point formats by directly operating on integers.
\item Mathematically figuring out the main cause of the overflow problem which is a serious issue of BP under integer-only arithmetic, and providing a solution for it.
\item A light and independent open source proof-of-concept C++ framework for PocketNN.
\end{itemize}

\section{Related Work}
Previous works on integer-only DNNs can be classified into two categories. The first category corresponds to quantizing inference only. BinaryConnect \cite{courbariaux2015binaryconnect}, Courbariaux et al. \cite{courbariaux2016binarized}, XNOR-net \cite{rastegari2016xnor}, TTQ \cite{zhu2016trained}, Benoit Jacob et al. \cite{jacob2018quantization}, and Ron Banner et al. \cite{banner2018scalable} belong to this category. BinaryConnect \cite{courbariaux2015binaryconnect}, XNOR-Net \cite{rastegari2016xnor} and TTQ \cite{zhu2016trained} push the quantization to the extreme that first two quantizes weights into 1 bit (i.e., binarization) and the third quantizes weights into 2 bits. Since 1 or 2 bits are too few for training, they keep using high precision floating-point real numbers for training and weight update. Then the real-valued weights are quantized for inference. The purpose of Benoit Jacob et al. \cite{jacob2018quantization} is different from other projects. It aims to find how to perform quantization aware training (QAT) wisely to achieve better accuracy than post training quantization (PTQ). It belongs to this category because it also uses floating-point real numbers for weights and biases. Ron Banner et al. \cite{banner2018scalable} showed 8-bit is enough for quantization. They also used floating-point numbers for operations on gradients.

The second category corresponds to quantizing both training and inference. DoReFa-Net \cite{zhou2016dorefa}, FxpNet \cite{chen2017fxpnet}, WAGE \cite{wu2018training}, and NITI \cite{wang2020niti} belong to this category. DoReFa-Net \cite{zhou2016dorefa} internally keeps track of floating-point weights for update and then quantizes them into fixed-point numbers. FxpNet \cite{chen2017fxpnet} tried further to use only fixed-point numbers and introduced integer batch normalization and fixed-point ADAM. However, it could not completely remove floating-point real numbers in their algorithm. WAGE \cite{wu2018training} and NITI \cite{wang2020niti} pursued the same goal. It should be noted that research projects in this category either use complex quantization algorithms or their own customized fixed-point notations to avoid floating-point real numbers. WAGE \cite{wu2018training} focused on how to quantize parameters via mapping, shifting and stochastic rounding. FxpNet \cite{chen2017fxpnet} and NITI \cite{wang2020niti} used two different integers $M$ and $E$, $M$ for mantissa and $E$ for exponent, to express a single real number $M \cdot 2^E$. Essentially, this approach is using two integers to simulate one floating-point real number of IEEE Standard for Floating-Point Arithmetic (IEEE 754) format \cite{ieee754}. This and other kinds of customized fixed-point formats are not universally agreed and therefore less compatible.

PocketNN trains and runs DNNs with only standard type integers without any floating-point or customized fixed-point formats. It directly operates on integers so that quantization is naturally done. Therefore PocketNN does not use explicit quantization algorithms. These characteristics make PocketNN simple, easily understandable, and highly compatible. As it will be explained in detail in the following section, these were made possible by direct feedback alignment (DFA) \cite{nokland2016direct} and self-designed pocket activation functions. To the best of our knowledge, this is the first research paper of training DNNs entirely via DFA with integer-only arithmetic. Although there is a very short web article with five sentences and one chart mentioning 4-bit quantization with DFA \cite{widialaksono2020} but there is no paper or code associated with it. Han et al. \cite{han2019direct} mentioned binarized DFA, yet their main focus was to amalgamate DFA at the end of backpropagation (BP) in convolutional neural network (CNN) models.

From an engineering point of view, it must be mentioned that PocketNN is fully written in pure C++ for maximum compatibility. It does not use any library, not even C++ STL, as popular low-end device families such as Arduino do not fully support C++ STL. Compared to most of the related work that uses Python and therefore need to be carefully ported to C++ before they can be tested on low-end devices, this is a unique feature of PocketNN that leads to higher compatibility.

\section{Methods and Algorithms}
In this section, all vectors are assumed to be row vectors. It is natural for a C++ framework because two-dimensional memory allocation in C++ can be easily interpretated as a set of row vectors.

\subsection{Integer backpropagation (BP) leads to overflow}
Backpropagation (BP) \cite{rumelhart1986learning} is a de facto standard DNN training algorithm. In BP, gradients for updating weight parameters are calculated by propagating the error backward from the output layer to the input layer (hence ``backpropagation''). BP can easily suffer from overflow problem in integer-only DNN training. Integer-only arithmetic is more vulnerable to this overflow problem because bits in integer format must be used to increase the absolute size of values while bits in floating-point format can be used to finely divide a small range.

Consider a row-wise input vector $\mathbf{x} \in \mathbb{R}^{1 \times d_0}$. Typical fully connected neural networks with $n$ layers can be formulated as:
\begin{eqnarray}
\mathbf{h}^{[1]} & = & \mathbf{x} W^{[1]} + \mathbf{b}^{[1]}\\
\mathbf{h}^{[k]} & = & \mathbf{a}^{[k-1]} W^{[k]} + \mathbf{b}^{[k]} \quad \textrm{(for $2 \leq k \leq n$)}\\
\mathbf{a}^{[k]} & = & \mathtt{actv}^{[k]}( \mathbf{h}^{[k]} )\\
\hat{\mathbf{y}} & = & \mathbf{a}^{[n]}
\end{eqnarray}
where $\mathbf{x} \in \mathbb{R}^{1 \times d_0}$ is the input vector; $W^{[k]} \in \mathbb{R}^{d_{k-1} \times d_k}$ is the weight matrix of the $k$th layer; $\mathbf{b}^{[k]}, \mathbf{h}^{[k]}, \mathbf{a}^{[k]} \in \mathbb{R}^{1 \times d_k}$ are the bias, summation of products and the bias, and output vector of the $k$th layer; $\mathbf{a}^{[n]}, \hat{\mathbf{y}} \in \mathbb{R}^{1 \times d_n} = \mathbb{R}^{1 \times 1}$ are the final output of the network; and $\mathtt{actv}^{[k]}(\cdot)$ is an activation function for the $k$th layer. Once $\hat{\mathbf{y}}$ is calculated, we can calculate the loss $J$ with the ground truth vector $\mathbf{y}$. Note that $d_n = 1$ and therefore both $\hat{\mathbf{y}}$ and $\mathbf{y}$ are scalars. For L2 loss, $J = {(\hat{\mathbf{y}} - \mathbf{y})^2} / 2$ is also a scalar.

For BP, let's define $\mathbf{\delta}_{\mathrm{BP}}^{[k]}$ (often called ``delta'') such that
\begin{eqnarray}
\mathbf{\delta}_{\mathrm{BP}}^{[k]} & \eqdef & \frac{\partial J}{\partial \mathbf{h}^{[k]}} \quad (\mathbf{\delta}_{\mathrm{BP}}^{[k]} \in \mathbb{R}^{1 \times d_k})
\end{eqnarray}
Once $\mathbf{\delta}_{\mathrm{BP}}^{[n]}$ is calculated, $\mathbf{\delta}_{\mathrm{BP}}^{[k]}$ for $1 \leq k \leq n-1$ can be recursively calculated. For L2 loss,
\begin{eqnarray}
\mathbf{\delta}_{\mathrm{BP}}^{[n]} & = & (\hat{\mathbf{y}} - \mathbf{y}) \odot \mathtt{actv}'^{[n]}(\mathbf{h}^{[n]})\\
&&(\hat{\mathbf{y}}, \mathbf{y}, \mathbf{h}^{[n]} \in \mathbb{R}^{1 \times 1}) \nonumber\\
\mathbf{\delta}_{\mathrm{BP}}^{[k]} & = & \mathbf{\delta}_{\mathrm{BP}}^{[k+1]} {W^{[k+1]}}^T \odot \mathtt{actv}'^{[k]}(\mathbf{h}^{[k]})\\
&&\textrm{(for $1 \leq k \leq n-1$)} \nonumber
\end{eqnarray}

With all $\mathbf{\delta}_{\mathrm{BP}}^{[k]}$s prepared, gradients of the loss function with respect to the weights and biases can be calculated. The calculated gradients can be multiplied by a learning rate to update weights and biases.
\begin{eqnarray}
\frac{\partial J}{\partial W^{[k]}} & = & {\mathbf{a}^{[k-1]}}^T \mathbf{\delta}_{\mathrm{BP}}^{[k]}\\
\frac{\partial J}{\partial \mathbf{b}^{[k]}} & = & \mathbf{\delta}_{\mathrm{BP}}^{[k]}
\end{eqnarray}

\begin{figure}[h]
\centering
\includegraphics[width=\linewidth]{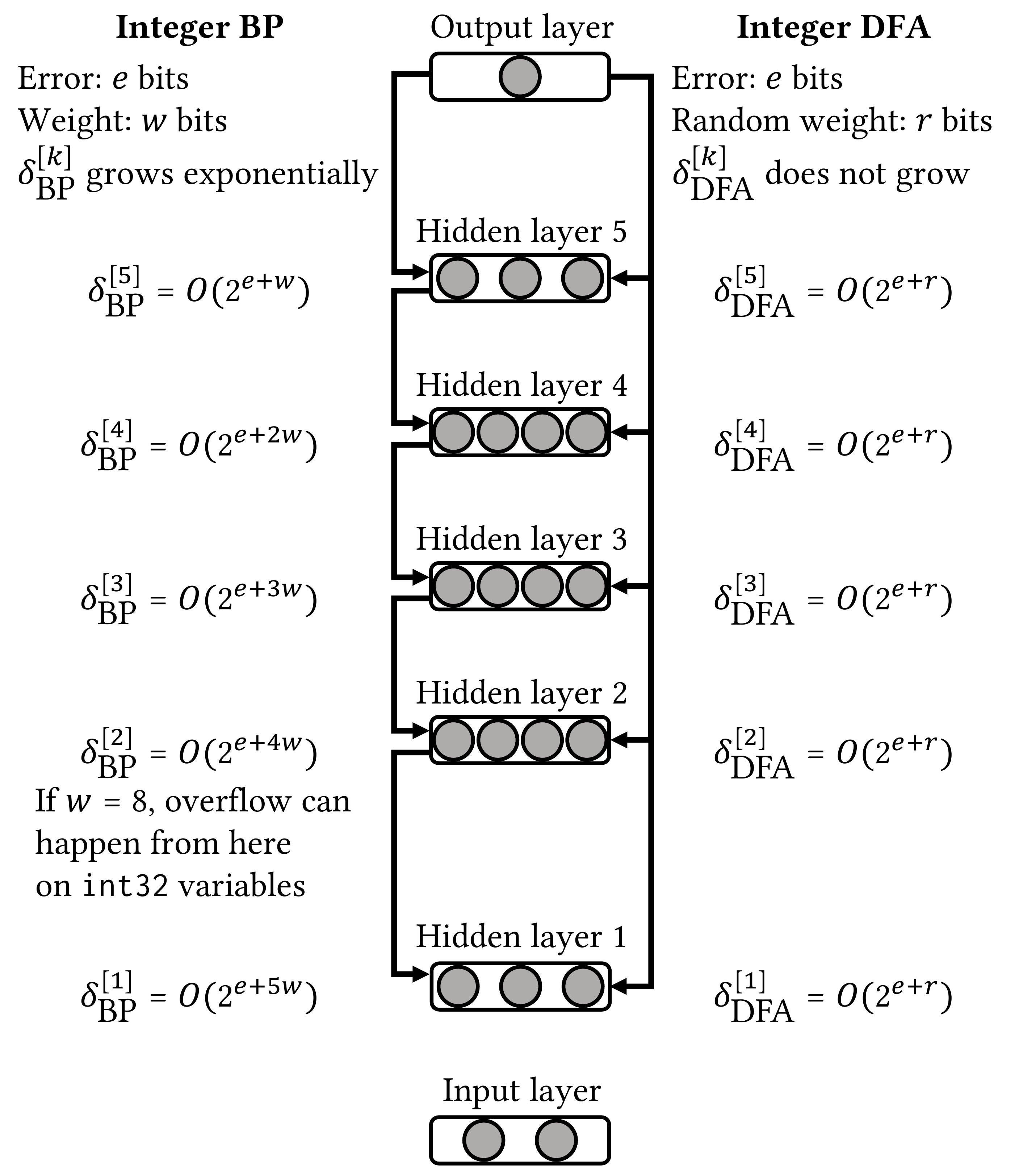}
\caption{Deltas of DFA and BP are shown on the sample fully connected neural network with five hidden layers. While upper bounds of DFA deltas remain same regardless of layers' depths, upper bounds of BP deltas increase exponentially layer by layer so that eventually integer overflow may occur. In the sample DNN, overflow can occur in hidden layers 1 and 2.}\label{fig:fc_overflow}
\Description{The image shows a sample fully connected neural network with five hidden layers. And Deltas of DFA and BP are shown on it. Upper bounds of DFA deltas remain same regardless of layers' depths. However, upper bounds of BP deltas increase exponentially layer by layer and eventually may cause integer overflow. In the sample network, overflow can occur in hidden layers 1 and 2.}
\end{figure}

Overflow can and very often occur in BP with integer-only arithmetic because deltas are recursively multiplied. Consider equation (7) again:
\begin{eqnarray*}
\mathbf{\delta}_{\mathrm{BP}}^{[k]} & = & \mathbf{\delta}_{\mathrm{BP}}^{[k+1]} {W^{[k+1]}}^T \odot \mathtt{actv}'^{[k]}(\mathbf{h}^{[k]})\\
&&\textrm{(for $1 \leq k \leq n-1$)}
\end{eqnarray*}
For the sake of simplicity, consider the situation where $\mathtt{actv}'^{[\cdot]}(\mathbf{h}^{[\cdot]})$ is not too small. If $\mathbf{\delta}_{\mathrm{BP}}^{[k+1]}$ requires $e$ bits and ${W^{[k+1]}}^T$ requires $w$ bits to represent their elements, their elements' upper bounds become $O(2^e)$ and $O(2^w)$, respectively. Then the upper bounds of $\mathbf{\delta}_{\mathrm{BP}}^{[k]}$, $\mathbf{\delta}_{\mathrm{BP}}^{[k-1]}$, $\mathbf{\delta}_{\mathrm{BP}}^{[k-2]}$, and generally $\mathbf{\delta}_{\mathrm{BP}}^{[k-p]}$, grow exponentially to $O(2^{e+w})$, $O(2^{e+2w})$, $O(2^{e+3w})$, and $O(2^{e + (p+1)w})$, respectively. Therefore eventually overflow may occur. Figure~\ref{fig:fc_overflow} describes this with a sample DNN. If $w = 8$ (\texttt{int8}), upper bounds for deltas will exceed 32 bits (\texttt{int32}) from the fourth hidden layer from the output and overflow can occur under 32-bit integer arithmetic. In practice, overflow occurs quite often in integer BP. Indeed, FxpNet \cite{chen2017fxpnet} and NITI \cite{wang2020niti} explicitly mentioned about overflow.

\subsection{Integer DFA can prevent overflow}
Direct feedback alignment (DFA) \cite{nokland2016direct} is a new emerging DNN training algorithm. DFA trains hidden layers independently from other layers by propagating error directly from the output layer to each individual hidden layer via fixed random feedback matrices. Thanks to this characteristic, DFA can be designed to be safe from overflow under integer-only arithmetic.

Instead of weights, appropriately sized random matrices $R^{[k]}$ are used to define deltas for DFA such that
\begin{eqnarray}
\mathbf{\delta}_{\mathrm{DFA}}^{[k]} & \eqdef & ((\hat{\mathbf{y}} - \mathbf{y}) R^{[k]}) \odot \mathtt{actv}'^{[k]}(\mathbf{h}^{[k]})\\
&&\textrm{(for $1 \leq k \leq n$)}\nonumber
\end{eqnarray}

It is important that $\mathbf{\delta}_{\mathrm{DFA}}^{[k]}$ is independent from $\mathbf{\delta}_{\mathrm{DFA}}$ of other layers (e.g., $\mathbf{\delta}_{\mathrm{DFA}}^{[k+1]}$). If upper bounds of $\hat{\mathbf{y}} - \mathbf{y}$ and $R^{[k]}$ are $O(2^e)$ and $O(2^r)$, respectively, then the maximum values of $\mathbf{\delta}_{\mathrm{DFA}}^{[k]}$ are bounded by $O(2^{e+r})$ for all $k$. For example, if $\hat{\mathbf{y}} - \mathbf{y}$ and $R^{[k]}$ use \texttt{int8} type variables, $O(2^{e+r})$ becomes $O(2^{16})$. Therefore \texttt{int32} will be more than enough to hold the values of $\mathbf{\delta}_{\mathrm{DFA}}^{[k]}$ for all $k$. Figure~\ref{fig:fc_overflow} describes this with graphical representation. An exceptional situation can occur if there are more than $O(2^{16})$ nodes in one layer because theoretically the maximum value can exceed \texttt{int32} in this case. However, practically there is no such DNN design, especially for low-end devices.

\subsection{Pocket activation functions}
\begin{table}
\caption{Mathematical formulae of pocket activation functions and their floating-point counterparts. PocketSigmoid and PocketTanh are approximations of rescaled sigmoid and tanh, respectively. PocketReLU8bit clamps input values between 0 and 127.}
\label{tab:actv_approx}
\setlength{\tabcolsep}{3pt}
\begin{tabular}{crlc}
\toprule
Activation Function & \multicolumn{2}{c}{Formula} & Range\\
\midrule
Sigmoid & \multicolumn{2}{c}{${e^x}/({e^x+1})$} & $(0, 1)$\\
$128 \cdot \mathrm{Sigmoid}(x/32)$ & \multicolumn{2}{c}{$128{e^{x/32}}/({e^{x/32}+1})$} & $(0, 128)$\\
PocketSigmoid (8 bit) & $1$ & ($x \leq -128$) & $[1, 127]$\\
& $x/8 + 20$  & ($-128 < x \leq -75$) &\\
& $x/2 + 48$  & ($-75 < x \leq -32$)  &\\
& $x + 64$    & ($-32 < x \leq 31$)   &\\
& $x/2 + 80$  & ($31 < x \leq 74$)    &\\
& $x/8 + 108$ & ($74 < x \leq 127$)   &\\
& $127$       & ($127 < x$)           &\\
\cmidrule{1-4}
Tanh & \multicolumn{2}{c}{$(e^{2x}-1)/(e^{2x}+1)$} & $(-1, 1)$\\
$128 \cdot \mathrm{Tanh}(x/64)$ & \multicolumn{2}{c}{$128(e^{2x/64}-1)/(e^{2x/64}+1)$} & $(-128, 128)$\\
PocketTanh (8 bit) & $-127$ & ($x \leq -128$) & $[-127, 127]$\\
& $x/4 - 88$ & ($-128 < x \leq -75$) &\\
& $x - 32$   & ($-75 < x \leq -32$)  &\\
& $2x$       & ($-32 < x \leq 31$)   &\\
& $x + 32$   & ($ 31 < x \leq 74$)   &\\
& $x/4 + 88$ & ($ 74 < x \leq 127$)  &\\
& $127$      & ($ 127 < x$)          &\\
\cmidrule{1-4}
ReLU6 & \multicolumn{2}{c}{min(max$(0, x), 6$)} & $[0, 6]$\\
PocketReLU8bit & \multicolumn{2}{c}{$\mathrm{min}(\mathrm{max}(0, x), 127)$} & $[0, 127]$\\
\bottomrule
\end{tabular}
\end{table}

\begin{figure}[h]
\centering
\includegraphics[width=\linewidth]{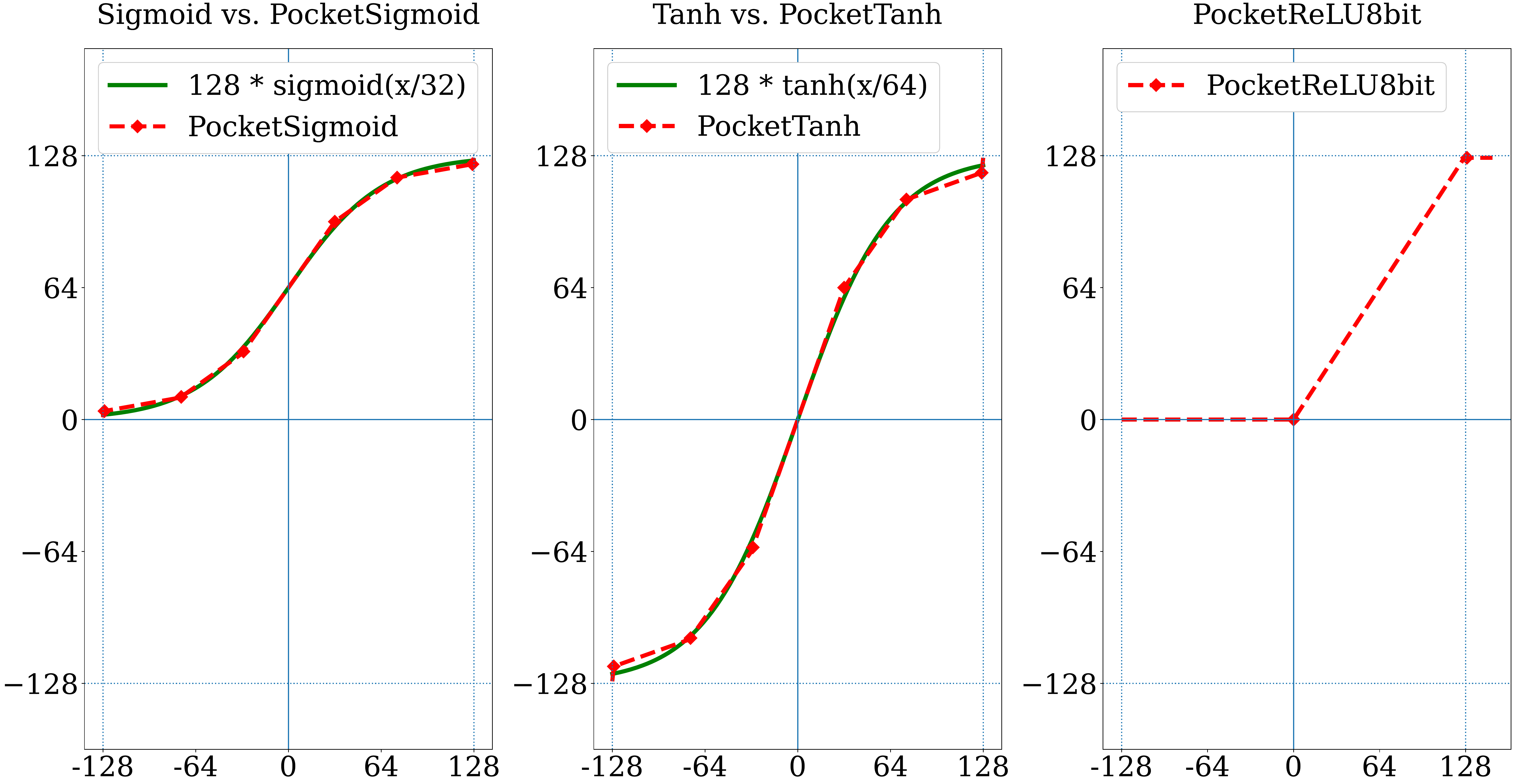}
\caption{Pocket activation functions. PocketSigmoid and PocketTanh are compared to sigmoid and tanh, the functions they linearly approximate, respectively. PocketReLU8bit is shown alone because it is not an approximation of another function.}\label{fig:pocket-activations}
\Description{Pocket activation functions are shown in the figure. PocketSigmoid is compared to sigmoid in the leftmost chart; PocketTanh is compared to tanh in the chart in the middle; PocketReLU8bit is shown alone in the rightmost chart because it is not an approximation of another function.}
\end{figure}

Activation functions are another main obstacle for the training and inference of DNNs with integer-only arithmetic. Logistic function (sigmoid), hyperbolic tangent (tanh) and rectified linear unit (ReLU) are popular activation functions. For sigmoid and tanh, the presence of the base of the natural logarithm $e$ in their formulae and their small ranges become problematic in integer-only arithmetic. Logistic function is $e^x/(e^x+1)$ and its return values are in the range between 0 and 1. It is hard to calculate $e^x$ with enough accuracy under integer-only arithmetic. Moreover, the return value will always be 0 because $e^x < e^x+1$ and C++'s signed integer division truncates towards zero. The situation is similar for tanh which is $(e^{2x}-1)/(e^{2x}+1)$. Therefore they need to be approximated and rescaled for integer-only arithmetic.

PocketNN provides a family of activation functions for integer-only DNNs. They include PocketSigmoid, PocketTanh, PocketReLU8bit, and are called pocket activations. Their mathematical formulae and graphs are shown in Table~\ref{tab:actv_approx} and Figure~\ref{fig:pocket-activations}. They are piecewise linear approximations of popular activation functions designed to take 8-bit input values and generate 8-bit output values to ensure consistency of behavior. If there is no such restriction, outputs from activation functions can explode or shrink to zero, causing unexpected behaviors such as integer overflow and vanishing gradients.

Piecewise linear approximation ensures simple and fast calculation under integer arithmetic. Inspired by Amin et al. \cite{amin1997piecewise}, PocketSigmoid and PocketTanh are designed to approximate the rescaled sigmoid and tanh such that -1, 0 and 1 in sigmoid and tanh's ranges are mapped to -127, 0 and 127 in pocket activations. So the ranges of PocketSigmoid and PocketTanh become $[1, 127]$ and $[-127, 127]$, respectively. It should be noted that PocketSigmoid never reaches 0 because sigmoid never reaches 0. The range of PocketTanh is not $[-128, 127]$ but $[-127, 127]$ to ensure symmetry.

Original $\tanh (x) = (e^{2x}-1)/(e^{2x}+1)$ equals to $2 \sigma (2x) - 1$ where $\sigma(x) = e^x/(e^{x} + 1)$ is the sigmoid function. This relationship does not hold between PocketTanh and PocketSigmoid. While PocketSigmoid approximates $128 \sigma(x/32)$, PocketTanh is approximates $128 \tanh (x/64) = 2 \cdot 128 \sigma (x/32) - 128$. Using $x/64$ instead of $x/32$ for PocketTanh achieved better performance because it made the output values more evenly distributed across its range $[-127, 127]$.

PocketReLU8bit is the PocketNN version of ReLU6 \cite{krizhevsky2010convolutional}. Just as ReLU6 is ReLU with 6 as its upper bound, PocketReLU8bit is ReLU with 127, the maximum value of \texttt{int8}, as its upper bound. PocketReLU8bit is used instead of general ReLU to prevent overflow. It should be noted that PocketTanh and PocketSigmoid usually work better than PocketReLU8bit with DFA. The original DFA paper \cite{nokland2016direct} also used tanh rather than ReLU.

\subsection{Other details}
Sum of squared errors (SSE) is the default loss function of PocketNN for classification tasks. It is because SSE can be accurately calculated by integer-only arithmetic. Although the combination of softmax and cross entropy loss is a popular approach for classification tasks, it is hard to calculate natural logarithm and exponential function $e^x$ with sufficient accuracy with integer-only arithmetic. WAGE \cite{wu2018training} also used SSE due to the same reason.

Proper fractions (fractions whose absolute values are strictly less than 1) are inevitably involved in learning rates and activation gradients. Learning rates are often below 1, and the slopes of sigmoid, tanh and ReLU are $(0, 0.25]$, $(0, 1]$ and $[0, 1]$, respectively. The situation is similar for pocket activations because their slopes are between 0 and 2. Therefore it is crucial to implement proper fractions with integer-only arithmetic.

PocketNN's simple solution is dividing by integers. For example, if learning rate is 0.01, PocketNN just divides weight updates by 100. Similarly, PocketSigmoid's $x/2 + 80$ for $31 < x \leq 74$ is calculated by dividing $x$ by 2 instead of multiplying 0.5. If an original slope is 0, multiplying 0 and dividing by \texttt{INT\_MAX} are both feasible because all values are below \texttt{INT\_MAX} in PocketNN.

\subsection{PocketNN code snippet}
PocketNN aims to be a research tool for tinyML. Therefore its C++ code and API are as equally important as algorithms. A fundamental building block of PocketNN API is \texttt{pktmat} class which stands for Pocket Matrix. It implements matrices via dynamic memory allocation and supports useful operations such as matrix arithmetic, rotations, slicing and random sampling. Activation functions and layers are implemented as \texttt{pktactv} and \texttt{pktlayer} classes using \texttt{pktmat} class.

Figure~\ref{fig:pocketnn-code} shows a sample PocketNN code snippet for loading MNIST images, constructing, training and running a fully connected neural network, and printing the output. Full PocketNN code and more examples can be found on its online repository. \footnote{\urlstyle{tt}\url{https://github.com/jaewoosong/pocketnn}}

\begin{figure}[h]
\centering
\begin{framed}
\begin{Verbatim}[baselinestretch=0.6]
// data loader
int numSamples = 60000;
pktmat trainLabels;
pktmat trainImages;
loadMnistLabels(trainLabels, numSamples, true);
loadMnistImages(trainImages, numSamples, true);

// numbers for the DNN
int numClasses = 10;
int mnistRows = 28;
int mnistCols = 28;
const int dim1 = 100;
const int dim2 = 50;

// DNN setup
pktactv::Actv a = pktactv::Actv::pocket_tanh;
pktfc fc1(mnistRows * mnistCols, dim1);
pktfc fc2(dim1, dim2);
pktfc fcLast(dim2, numClasses);

// connect layers
fc1.useDfa(true).setActv(a).setNextLayer(fc2);
fc2.useDfa(true).setActv(a).setNextLayer(fcLast);
fcLast.useDfa(true).setActv(a);

// (some code omitted for brevity)
// forward and backward passes
fc1.forward(mnistTestImages);
batchL2LossDelta(deltas, targets, fcLast.mOutput);
fcLast.backward(deltas, learningRateInverse);

// print output values
fcLast.printOutput();
\end{Verbatim}
\end{framed}
\caption{Sample C++ code snippet of PocketNN API for training a DNN with the MNIST dataset.}
\label{fig:pocketnn-code}
\Description{This is a sample C++ code snippet of PocketNN API. The code shows how to train a PocketNN DNN model with the MNIST dataset.}
\end{figure}

\section{Results}
PocketNN was compared to floating-point BP on two well-known image datasets, the MNIST \cite{lecun1998gradient} and Fashion-MNIST \cite{fashionmnist2017}. The MNIST and Fashion-MNIST datasets consist of grayscale handwritten digit and fashion item images, respectively. Both have 60,000 training and 10,000 validation images. Each image consists of $784 (28 \times 28)$ pixels.

\begin{figure}[h]
\centering
\includegraphics[width=\linewidth]{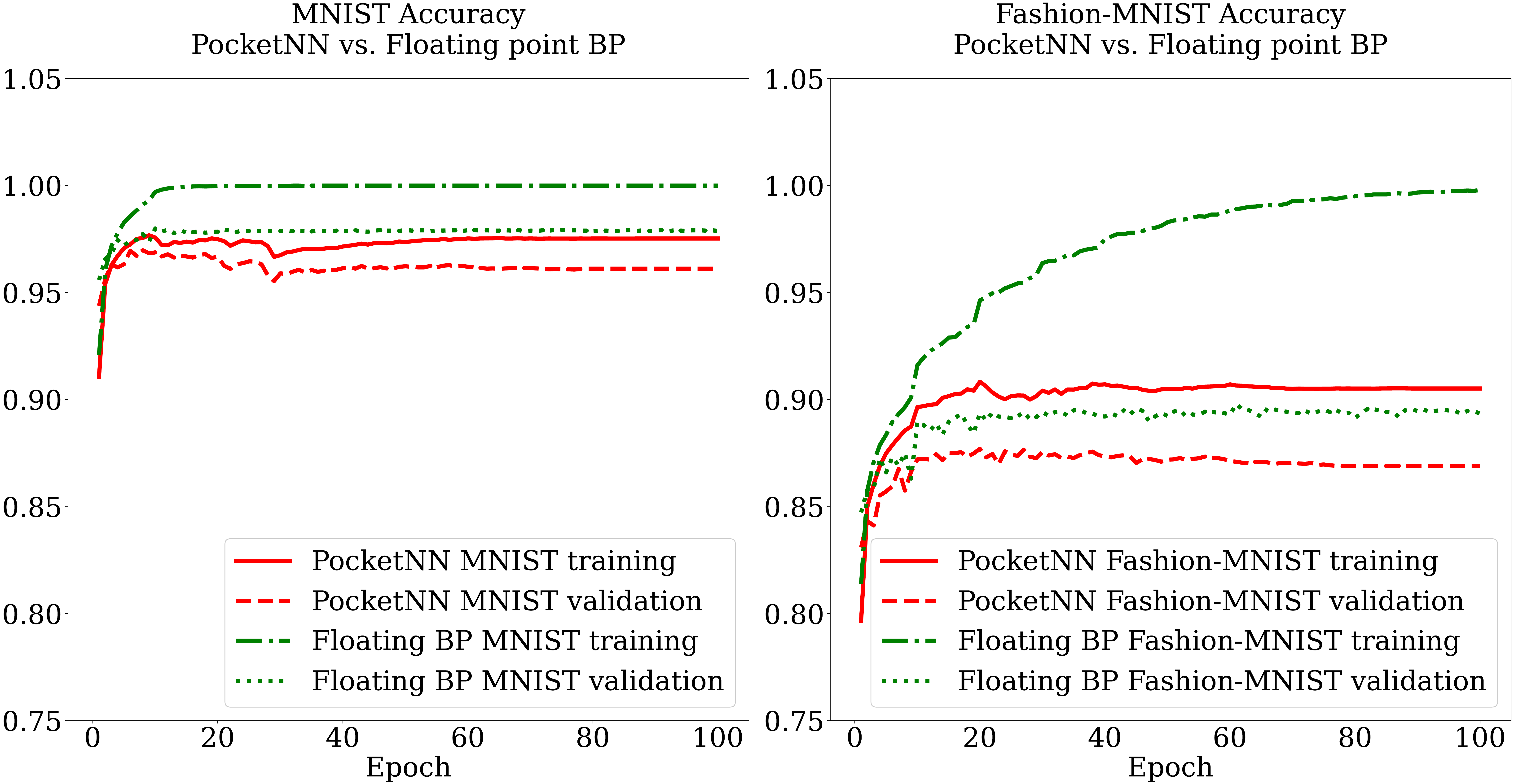}
\caption{Comparison of PocketNN and floating-point BP with SGD on MNIST and Fashion-MNIST datasets.}\label{fig:results-chart}
\Description{TBD}
\end{figure}

Simple fully-connected neural network models were constructed to compare  PocketNN with floating-point BP. For both datasets, input and output layers consisted of 784 (28 x 28) and 10 (10 categories) nodes, respectively. For the MNIST dataset, there were 2 hidden layers of 100 and 50 nodes; for the Fashion-MNIST dataset, 3 hidden layers of 200, 100 and 50 nodes were used. PocketNN models used PocketTanh and sum of squared error (SSE) as activation and loss functions, respectively, and was trained by integer DFA. The floating-point BP models used tanh as an activation function, sparse categorical cross entropy (sparse CCE) as a loss function and was trained by stochastic gradient descent (SGD). PocketNN models' weights were all initialized as zeros and the floating-point BP models' weights were initialized via Glorot uniform initialization \cite{glorot2010understanding}. Biases were initialized as zeros in both cases. Training was performed for 100 epochs with batch size 20. Learning rates started from 0.001 for PocketNN and 0.1 for floating-point BP because those values gave better results to each algorithm. The learning rates were halved by after each 10 epochs. Figure~\ref{fig:results-chart} shows the training and validation accuracy curves for 100 epochs.

\begin{table}
\caption{Comparison of PocketNN and floating-point BP with SGD on MNIST and Fashion-MNIST datasets. PocketNN used PocketTanh and SSE as activation and loss functions, and was trained via integer DFA. Floating-point BP used tanh and sparse CCE as activation and loss functions, and was trained via floating-point SGD.}
\label{tab:result}
\setlength\tabcolsep{2.5pt}
\begin{tabular}{cccc}
\toprule
\makecell{Dataset \& \\DNN Structure} & PocketNN & Floating-point BP & \makecell{Accuracy\\degradation}\\
\midrule
\makecell{MNIST\\784-100-50-10} & 96.98\% & 98.0\% & 1.02\%p\\
\cmidrule{1-4}
\makecell{Fashion-MNIST\\784-200-100-50-10} & 87.7\% & 89.79\% & 2.09\%p\\
\bottomrule
\end{tabular}
\end{table}

Table~\ref{tab:result} shows the best validation accuracies of PocketNN and floating-point BP models. PocketNN models achieved validation accuracies of 96.98\% for MNIST and 87.7\% for Fashion-MNIST. The results were close to the results of floating-point BP models such that accuracy degradations were 1.02\%p and 2.09\%p, respectively.

\section{Discussion and Future Work}
PocketNN's performance should be compared to WAGE \cite{wu2018training} and NITI \cite{wang2020niti}. They are the only two related works which implemented their full algorithms using fixed-point formats only, albeit WAGE \cite{wu2018training} sometimes used floating-point softmax. Other previous works used floating-point arithmetic in significant parts of their algorithms. However, it is hard to directly compare PocketNN to them because PocketNN used fully connected neural networks (FCNNs) while WAGE \cite{wu2018training} and NITI \cite{wang2020niti} used convolutional neural networks (CNNs) for experiments. PocketNN targeted FCNNs because they are more fundamental and generalizable DNN structure. For example, CNNs are often transformed into FCNNs for faster calculation via im2col algorithm \cite{chellapilla2006high}.

Since direct comparison is not feasible, it will be a good measure to compare the differences between each approach's results with their floating-point BP counterparts' results. WAGE \cite{wu2018training} compared training curves of their models to the training curve of a vanilla CNN on the CIFAR-10 dataset \cite{krizhevsky2009learning} and their quantized models showed about 1\%p to 1.5\%p accuracy degradations. NITI \cite{wang2020niti} mentioned their quantized models achieved negligible accuracy degradation on the MNIST dataset \cite{lecun1998gradient} using LeNet structure \cite{lecun1995learning} and 2.8\% lower accuracy on the CIFAR-10 dataset \cite{krizhevsky2009learning} with VGG-13 structure \cite{simonyan2014very}. Meanwhile, it is worth noting that the original DFA paper \cite{nokland2016direct} reported that \textit{floating-point} DFA showed almost identical accuracies on MNIST dataset and 0.7\% to 2.3\% points lower accuracies on CIFAR-10 dataset compared to their BP equivalents.

Compared with WAGE \cite{wu2018training}, NITI \cite{wang2020niti}, and the original DFA paper \cite{nokland2016direct}, PocketNN achieved almost identical performance with much simpler integer-only training and inference algorithms which helps solving integer overflow problem and does not require any complex quantization schemes.

In future work, it will be important to apply PocketNN to other DNN architectures such as convolutional neural networks (CNNs), recurrent neural networks (RNNs) and residual network (ResNet) \cite{he2016deep} on other popular datasets such as CIFAR-10, CIFAR-100 and ImageNet.

\section{Conclusion}
PocketNN is an algorithm for the integer-only training and inference of deep neural networks, and also a proof-of-concept C++ framework for implementing that algorithm. It runs directly on integers without explicit complex quantization algorithms or customized fixed-point formats. It avoids integer overflow by using DFA instead of BP for training DNNs. A family of newly designed activation functions for integer-only DNNs are introduced as pocket activations. PocketNN achieved 96.98\% and 87.7\% accuracies on MNIST and Fashion-MNIST datasets, respectively, showing only marginal accuracy degradations compared to its floating-point BP counterparts. PocketNN is coded into a light open source C++ framework without any dependencies to ensure maximum compatibility and portability to low-end devices. It will be a useful tool for tinyML researchers and developers working on integer-only DNNs.

\begin{acks}
Thanks to Jae Boum Kim for his valuable suggestions for datasets.
\end{acks}

\bibliographystyle{ACM-Reference-Format}
\bibliography{pocketnn_bibfile}


\end{document}